\documentclass[journal,twoside]{IEEEtran}
\usepackage{graphicx}
\usepackage{multirow}
\usepackage{textcomp}
\usepackage{subcaption}
\usepackage{amsmath}
\usepackage{breqn}
\usepackage[usestackEOL]{stackengine}
\usepackage{mathtools}
\usepackage{lipsum}
\usepackage{pdflscape}
\usepackage{tabularx,booktabs, longtable}
\begin{document}

\title{Deep Learning for Neuroimaging-based Diagnosis and Rehabilitation of Autism Spectrum Disorder:\\A Review  }

\author{Marjane~Khodatars,
        Afshin~Shoeibi,
        Delaram~Sadeghi,
        Navid~Ghassemi,
        Mahboobeh~Jafari,
        Parisa~Moridian,
        Ali~Khadem,
        Roohallah~Alizadehsani,
        Assef~Zare,
        Yinan~Kong,
        Abbas~Khosravi,
        Saeid~Nahavandi,
        Sadiq~Hussain,
        ~U.~Rajendra~Acharya,
        Michael~Berk
\thanks{M. Khodatars and D. Sadeghi are with the Dept. of Medical Engineering, Mashhad Branch, Islamic Azad University, Mashhad, Iran.}
\thanks{A. Shoeibi and N. Ghassemi are with the Faculty of Electrical Engineering, FPGA Lab, K. N. Toosi University of Technology, Tehran, Iran, and the Computer Engineering Department, Ferdowsi University of Mashhad, Mashhad, Iran. (Corresponding author: Afshin Shoeibi, email: afshin.shoeibi@gmail.com).}% <-this % stops a space
\thanks{M. Jafari is with Electrical and Computer Engineering Faculty, Semnan University, Semnan, Iran.}
\thanks{P. Moridian is with the Faculty of Engineering, Science and Research Branch, Islamic Azad University, Tehran, Iran.}
\thanks{A. Khadem is with the Faculty of Electrical Engineering, K. N. Toosi University of Technology, Tehran, Iran. (Corresponding author: Ali Khadem, email: alikhadem@kntu.ac.ir).}
\thanks{R. Alizadehsani, A. Khosravi and S. Nahavandi. are with the Institute for Intelligent Systems Research and Innovation (IISRI), Deakin University, Victoria 3217, Australia.}
\thanks{A. Zare is with Faculty of Electrical Engineering, Gonabad Branch, Islamic Azad University, Gonabad, Iran.}
\thanks{Y. Kong is with the School of Engineering, Macquarie University, Sydney, 2109, Australia.}
\thanks{S. Hussain is with the Dibrugarh University, Assam, India, 786004.}
\thanks{U. R. Acharya is with the Dept. of Electronics and Computer Engineering, Ngee Ann Polytechnic, Singapore 599489, Singapore, the Dept. of Biomedical Informatics and Medical Engineering, Asia University, Taichung, Taiwan, and the Dept. of Biomedical Engineering, School of Science and Technology, Singapore University of Social Sciences, Singapore.}
\thanks{M. Berk is with the Deakin University, IMPACT - the Institute for Mental and Physical Health and Clinical Translation, School of Medicine, Barwon Health, Geelong, Australia, and the  Orygen, The National Centre of Excellence in Youth Mental Health, Centre for Youth Mental Health, Florey Institute for Neuroscience and Mental Health and the Department of Psychiatry, The University of Melbourne, Melbourne, Australia.}
% <-this % stops a space
% \thanks{Manuscript received April 19, 2005; revised August 26, 2015.}
}

%\markboth{IEEE Transactions on Cybernetics}{Khodatars \MakeLowercase{\textit{et al.}}: Deep Learning for Neuroimaging-based Diagnosis and Rehabilitation of Autism Spectrum Disorder}

\maketitle

\begin{abstract}
Accurate diagnosis of Autism Spectrum Disorder (ASD) followed by effective rehabilitation is essential for the management of this disorder. Artificial intelligence (AI) techniques can aid physicians to apply automatic diagnosis and rehabilitation procedures. AI techniques comprise traditional machine learning (ML) approaches and deep learning (DL) techniques. Conventional ML methods employ various feature extraction and classification techniques, but in DL, the process of feature extraction and classification is accomplished intelligently and integrally. DL methods for diagnosis of ASD have been focused on neuroimaging-based approaches. Neuroimaging techniques are non-invasive disease markers potentially useful for ASD diagnosis. Structural and functional neuroimaging techniques provide physicians substantial information about the structure (anatomy and structural connectivity) and function (activity and functional connectivity) of the brain. Due to the intricate structure and function of the brain, proposing optimum procedures for ASD diagnosis with neuroimaging data without exploiting powerful AI techniques like DL may be challenging. In this paper, studies conducted with the aid of DL networks to distinguish ASD are investigated. Rehabilitation tools provided for supporting ASD patients utilizing DL networks are also assessed. Finally, we will present important challenges in the automated detection and rehabilitation of ASD and propose some future works.
\end{abstract}


\IEEEpeerreviewmaketitle

\section{Introduction}

\IEEEPARstart{A}{SD} is a disorder of the nervous system that affects the brain and results in difficulties in speech, social interaction and communication deficits, repetitive behaviors, and delays in motor abilities \cite{one}. This disease can generally be distinguished with extant diagnostic protocols from the age of three years onwards. Autism influences many parts of the brain. This disorder also involves a genetic influence via the gene interactions or polymorphisms \cite{two,three}. One in 70 children worldwide is affected by autism. In 2018, the prevalence of ASD was estimated to occur in 168 out of 10,000 children in the United States, one of the highest prevalence rates worldwide. Autism is significantly more common in boys than in girls. In the United States, about 3.63 percent of boys aged 3 to 17 years have autism spectrum disorder, compared with approximately 1.25 percent of girls \cite{four}.

Diagnosing ASD is difficult because there is no pathophysiological marker, relying instead just on psychological criteria \cite{five}. Psychological tools can identify individual behaviors, levels of social interaction, and consequently facilitate early diagnosis. Behavioral evaluations embrace various instruments and questionnaires to assist the physicians to specify the particular type of delay in a child's development, including clinical observations, medical history, autism diagnosis instructions, and growth and intelligence tests \cite{six}.

Several investigations for the diagnosis of ASD have recently been conducted on neuroimaging data (structural and functional).

Analyzing anatomy and structural connections of brain areas with structural neuroimaging is an essential tool for studying structural disorders of the brain in ASD. The principal tools for structural brain imaging are magnetic resonance imaging (MRI) techniques \cite{seven,a8,a9}. Cerebral anatomy is investigated by structrul MRI (sMRI) images and anatomical connections are assesed by diffusion tensor imaging MRI (DTI-MR) \cite{a10}. Investigating the activity and functional connections of brain areas using functional neuroimaging can also be used for studying ASD. Brain functional diagnostic tools are older approaches than structural methods for studying ASD. The most basic modality of functional neuroimaging is electroencephalography (EEG), which records the electrical activity of the brain from the scalp with a high temporal resolution (in milliseconds order) \cite{a11}. Studies have shown that employing EEG signals to diagnose ASD have been useful \cite{a12,a13,a14}. Functional MRI (fMRI) is one of the most promising imaging modalities in functional brain disorders, used as task-based (T-fMRI) or resting‐state (rs-fMRI) \cite{a15,a16}. fMRI-based techniques have a high spatial resolution (in the order of millimeters) but a low temporal resolution due to slow response of the hemodynamic system of the brain as well as fMRI imaging time constraints and is not ideal for recording the fast dynamics of brain activities. In addition, these techniques have a high sensitivity to motion artifacts. It should be stressed that in consonance with studies, three less prevalent modalities of electrocorticography (ECoG) \cite{a17}, functional near-infrared spectroscopy (fNIRS) \cite{a18}, and Magnetoencephalography (MEG) \cite{a19} can also attain reasonable performance in ASD diagnosis. An appropriate approach is to utilize machine-learning techniques alongside functional and structural data to collaborate with physicians in the process of accurately assessing ASD. In the field of ASD, applying machine learning methods generally entail two categories of traditional methods \cite{a20} and DL methods \cite{a21}. As opposed to traditional methods, much less work has been done on DL methods to explore ASD or design rehabilitation tools.

This study reviews ASD assesment methods and patients' rehabilitation with DL networks. The outline of this paper is as follows. Section 2 is search strategy. Section 3 concisely presents the DL networks employed in the field of ASD. In section 4, existing computer-aided diagnosis systems (CADS) that use brain functional and structural data are reviewed. In section 5, DL-based rehabilitation tools for supporting ASD patients are introduced. Section 6 discusses the reviewed papers. Section 7 reveals the challenges of ASD diagnosis and rehabilitation with DL. Finally, the paper concludes and suggests future work in section 8.  
\section{Search Strategy}
\begin{figure}[t]
    \centering
    \includegraphics[width=3.5in ]{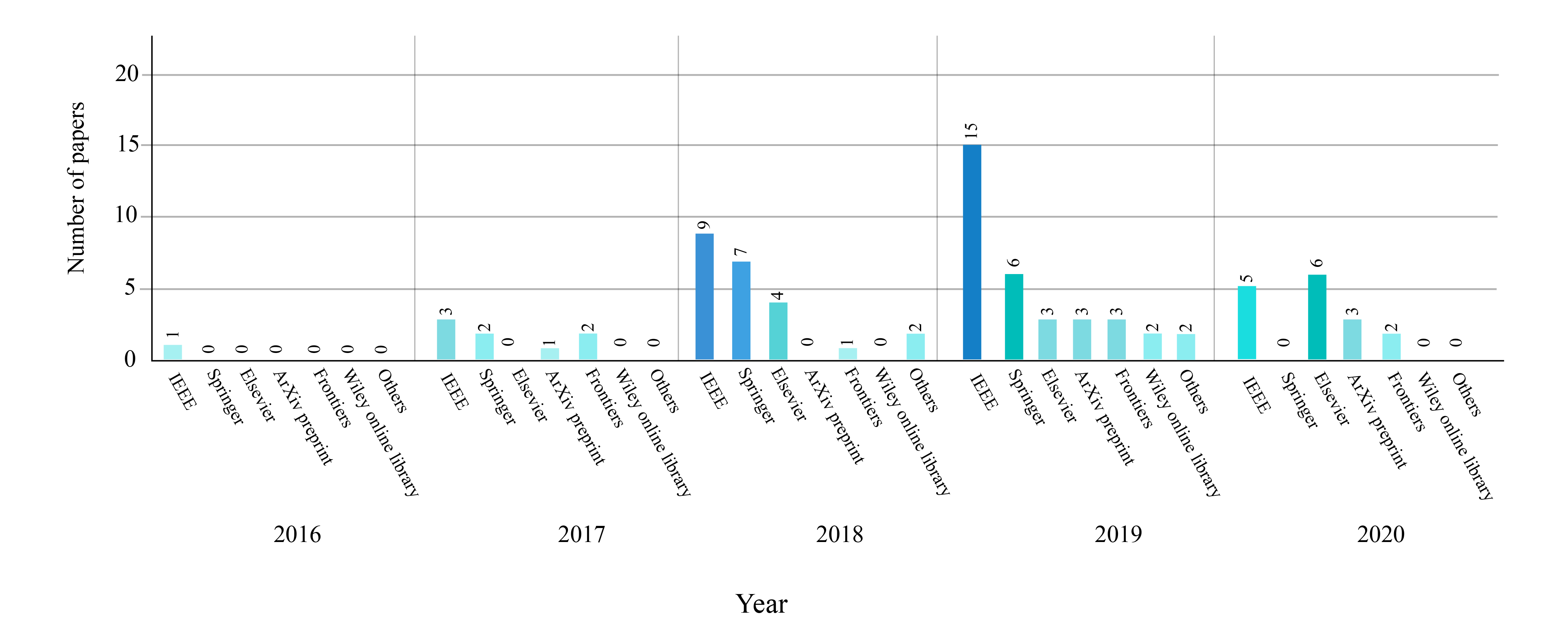}
    
    \caption{Number of papers published every year for ASD diagnosis and rehabilitation.}
    \label{fig:one}
\end{figure}

In this review, IEEE Xplore, ScienceDirect, SpringerLink, ACM, as well as other conferences or journals were used to acquire papers on ASD diagnosis and rehabilitation using DL methods. Further, the keywords "ASD", "Autism Spectrum Disorder" and "Deep Learning" were used to select the papers. The papers are analyzed till June 03th, 2020 by the authors (AK, SN). Figure \ref{fig:one} depicts the number of considered papers using DL methods for the automated detection and rehabilitation of ASD each year. 

\section{Deep Learning Techniques for ASD Diagnosis and Rehabilitation}
\begin{figure}[t]
    \centering
    \includegraphics[width=3.5in ]{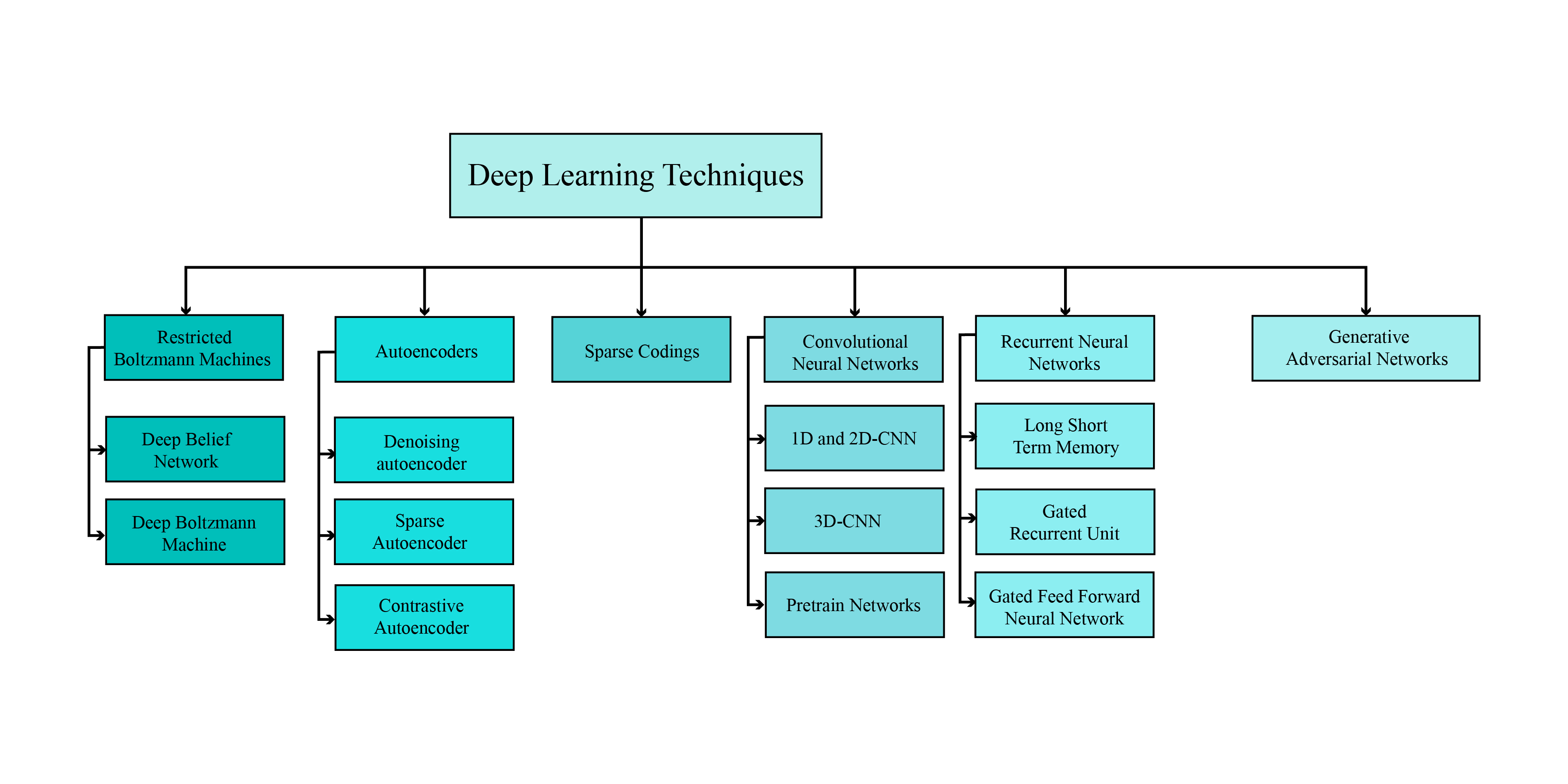}
    
    \caption{Illustration of various types of DL methods.}
    \label{fig:two}
\end{figure}

Nowadays, DL algorithms are used in many areas of medicine including structural and functional neuroimaging. The application of DL in neural imaging ranges from brain MR image segmentation \cite{a22}, to detection of brain lesions such as tumors \cite{a23}, diagnosis of brain functional disorders such as ASD \cite{a65}, and production of artificial structural or functional brain images \cite{a24}. Machine learning techniques are categorized into three fundamental categories of learning: supervised learning \cite{a25}, unsupervised learning \cite{a26}, and reinforcement learning \cite{a27}, and a variety of DL networks are provided for each type. So far, most studies applied to identify ASD using DL have been based on supervised or unsupervised approaches. Figure \ref{fig:two} illustrates generally employed types of DL networks with supervised or unsupervised learning to study ASD.

\section{CADS-based deep learning techniques for ASD diagnosis by neuroimaging data }

\begin{figure*}[ht]
    \centering
    \includegraphics[width=7in ]{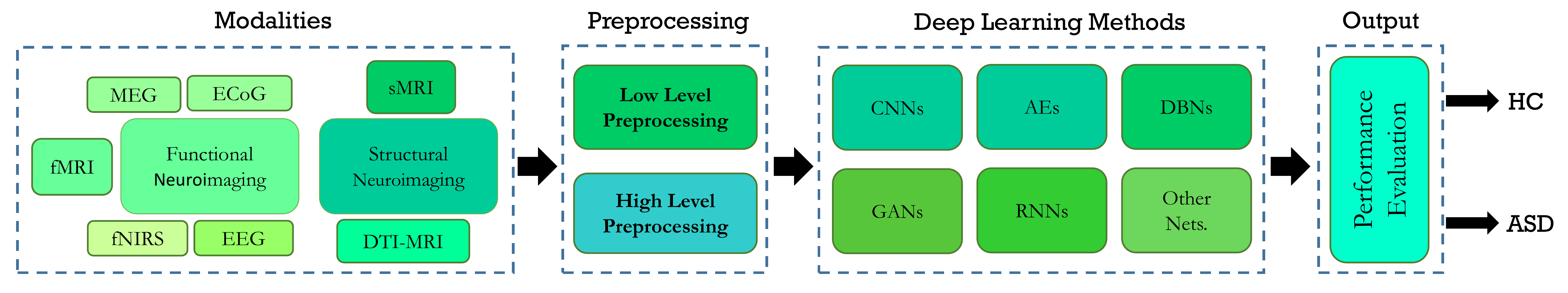}  
    
    \caption{Block diagram of CAD system using DL architecture for ASD detection.}
    \label{fig:int}
\end{figure*}

A traditional artificial intelligence (AI)-based CADS encompasses several stages of data acquisition, data pre-processing, feature extraction, and classification \cite{a28,a29,a30,a146}. In \cite{a31,a32,a33} existing traditional algorithms for diagnosing ASD have been reviewed. In contrast to traditional methods, in DL-based CADS, feature extraction, and classification are performed intelligently within the model. Also, due to the structure of DL networks, using large dataset to train DL networks and recognize intricate patterns in datasets is incumbent. The components of DL-based CADS for ASD detection are shown in Figure \ref{fig:int}. It can be noted from the figure that, large and free databases are first introduced to diagnose ASD. In the second step, various types of pre-processing techniques are used on functional and structural data to be scrutinized. Finally, the DL networks are applied on the preprocessed data.

\subsection{Neuroimaging ASD Datasets}
Datasets are the heart of any CADS development and the capability of CADS depends primarily on the affluence of the input data. To diagnose ASD, several brain functional and structural datasets are available. The most complete free dataset available is ABIDE \cite{a34} dataset with two subsets: ABIDE-I and ABIDE-II, which encompasses sMRI, rs-fMRI, and phenotypic data. ABIDE-I involves data from 17 international sites, yielding a total of 1112 datasets, including 539 individuals with ASD and 573 healthy individuals (ages 64-7). In accordance with HIPAA guidelines and 1000 FCP / INDI protocols, these data are anonymized.  In contrast, ABIDE-II contains data from 19 international sites, with a total of 1114 datasets from 521 individuals with ASD and 593 healthy individuals (ages 5-64). Also, preprocessed images of the ABIDE-I series called PCP \cite{a35} can be freely downloaded by the researchers. The second recently released ASD diagnostic database is called NDAR, which comprises various modalities, and more information is provided in \cite{a36}.
\subsection{Preprocessing Techniques}
Neuroimaging data (especially functional ones) is of relatively complicated structure, and if not pre-processed properly, it may affect the final diagnosis. Preprocessing of this data typically entails multiple common steps performed by different software as standard. Indeed, occasionally prepared pipelines are applied on the dataset to yield pre-processed data for future researches. In the following section, preprocessing steps are briefly explained for fMRI data.
\subsubsection{Standard (Low-level) fMRI preprocessing steps}
Low-level pre-processing of fMRI images normally has fixed number of steps applied on the data, and prepared toolboxes are usually used to reduce execution time and yield better accuracy. Some of these reputable toolboxes contain FMRIB software libraries (FSL) \cite{a37}, BET \cite{a38}, FreeSurfer \cite{a39}, and SPM \cite{a40}. Also, important and vital fMRI preprocessing incorporates brain extraction, spatial smoothing, temporal filtering, motion correction, slice timing correction, intensity normalization, and registration to standard atlas, which are summarized as follows:

\textsc{Brain extraction:} the goal is to remove the skull and cerebellum from the fMRI image and maintain the brain tissue \cite{a41,a42,a43}. 

\textsc{Spatial smoothing:} involves averaging the adjacent voxels signal. This process is persuasive on account of neighboring brain voxels being usually closely related in function and blood supply \cite{a41,a42,a43}. 

\textsc{Temporal filtering:} the aim is to eliminate unwanted components from the time series of voxels without impairing the signal of interest \cite{a41,a42,a43}.

\textsc{Realignment (Motion Correction):} During the fMRI test, people often move their heads. The objective of motion correction is to align all images to a reference image so that the coordinates and orientation of the voxels be identical in all fMRI volumetric images \cite{a41,a42,a43}. 

\textsc{Slice Timing Correction:} The purpose of modifying the slice time is to adjust the time series of the voxels so that all the voxels in each fMRI volume image have a common reference time. Usually, the corresponding time of the first slice recorded in each fMRI volume image is selected as the reference time \cite{a41,a42,a43}. 

\textsc{Intensity Normalization:} at this stage, the average intensity of fMRI signals are rescaled to compensate for global deviations within and between the recording sessions \cite{a41,a42,a43}. 

\textsc{Registration to a standard atlas:} The human brain entails hundreds of cortical and subcortical areas with various structures and functions, each of which is very time-consuming and complex to study. To overcome the problem, brain atlases are employed to partition brain images into a confined number of ROIs, following which the mean time series of each ROI can be extracted \cite{a44}. ABIDE datasets use a manifold of atlases, including Automated Anatomical Labeling (AAL) \cite{a45}, Eickhoff-Zilles (EZ) \cite{a46}, Harvard-Oxford (HO) \cite{a47}, Talaraich and Tournoux (TT) \cite{a48}, Dosenbach 160 \cite{a49}, Craddock 200 (CC200) \cite{a50} and Craddock 400 (CC400) \cite{a51} and more information is provided in \cite{a52}. Table \ref{tablelast} provides complete information on preprocessing tools, atlases, and some other preprocessing information.

\begin{figure*}[ht]
    \centering
    \includegraphics[width=7in ]{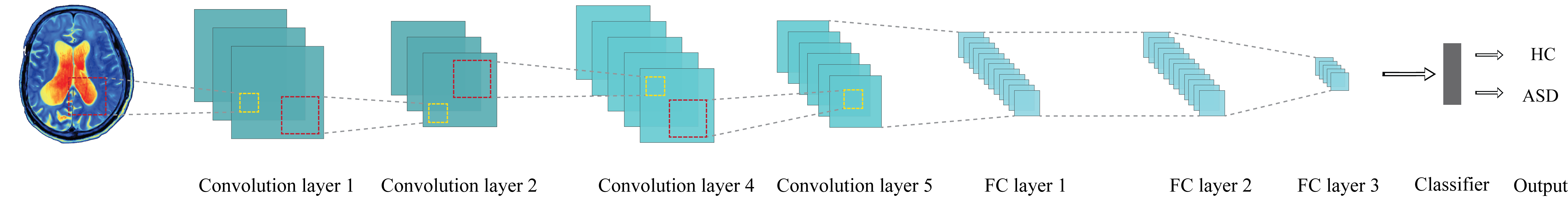}  
    
    \caption{Overall block diagram of a 2D-CNN used for ASD detection.}
    \label{fig:cnn2d}
\end{figure*}
\begin{figure*}[ht]
    \centering
    \includegraphics[width=6.5in ]{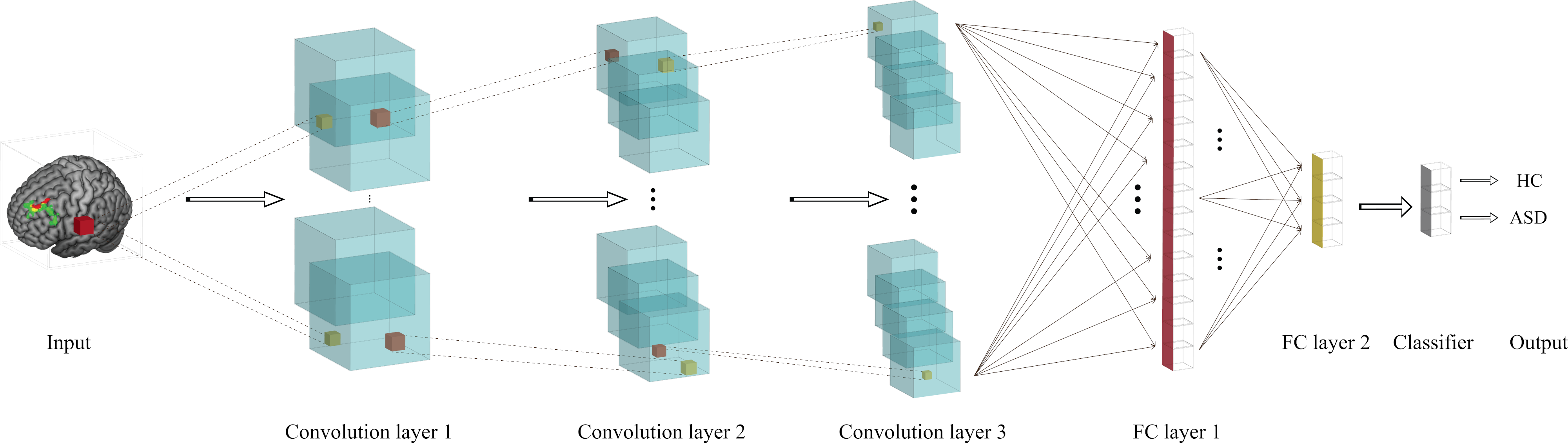}  
    
    \caption{Overall block diagram of a 3D-CNN used for ASD detection.}
    \label{fig:cnn3d}
\end{figure*}

\subsubsection{Pipeline Methods}
Pipelines present preprocessed images of ABIDE databases. They embrace generic pre-processing procedures. Employing pipelines, distinct methods can be compared with each other. In ABIDE datasets, pre-processing is performed by four pipeline techniques: neuroimaging analysis kit (NIAK) \cite{a53}, data processing assistant for rs- fMRI (DPARSF) \cite{a54}, the configurable pipeline for the analysis of connectomes (CPAC) \cite{a55}, or connectome computation system (CCS) \cite{a56}. The preprocessing steps carried out by the various pipelines are comparatively analogous. The chief differences are in the particular algorithms for each step, the software simulations, and the parameters applied. Details of each pipeline technique are provided in \cite{a52}. Table \ref{tablelast} demonstrates the pipeline techniques used in autism detection exploiting DL. 
\subsubsection{High-level preprocessing Steps}
High-level techniques for pre-processing brain data are important, and using them accompanying preliminary pre-processing methods can enhance the accuracy of ASD recognition. These methods are applied after the standard pre-processing of functional and structural brain data. These include sliding window (SW) \cite{a65}, data augmentation (DA) \cite{a68}, functional connectivity matrix (FCM) estimation \cite{a92,a93} and applying fast Fourier transformation (FFT) \cite{a78}. Furthermore, some of the researches utilized feature extraction \cite{a106} techniques and some also use feature selection methods. Precise information on reviewed studies is indicated in detail in Table \ref{tablelast}.

\subsection{Deep Neural Networks}
Deep learning in various medical applications, including the diagnosis of ASD, has become extremely popular in recent years. In this section of the paper, the types of Deep Learning networks used in ASD detection are examined, which include CNN, RNN, AE, DBN, CNN-RNN, and CNN-AE models.

\subsubsection{Convolutional Neural Networks (CNNs) }
In this section, the types of popular convolutional networks used in ASD diagnosis are surveyed. These networks involve 1D-CNN, 2D-CNN, 3D-CNN models, and a variety of pre-trained networks such as VGG.

\textsc{1D and 2D-CNN}

There are many spatial dependancies present in the data and it is difficult to extract these hidden signatures from the data. Convolution network uses a structure alike to convolution filters to extract these features properly and contribute to the knowledge that features should be processed taking into account spatial dependencies; so the number of network parameters are significantly reduced. The principal application of these networks is in image processing and due to the two-dimensional (2D) image inputs, convolution layers form 2D structures, which is why these networks are called 2D convolutional neural network (2D-CNN). By using another type of data, one-dimensional signals, the convolution layers' structure also resembles the data structure \cite{a57}. In convolution networks, assuming that various data sections do not require learning different filters, the number of parameters are markedly lessened and make it feasible to train these networks with smaller databases \cite{a21}. Figure \ref{fig:cnn2d} shows the block digram of 2D-CNN used for ASD detection.

\textsc{3D-CNN}

By transforming the data into three dimensions, the convolution network will also be altered to a three-dimensional format (Figure \ref{fig:cnn3d}). It should be noted that the manipulation of three dimensional CNN (3D-CNN) networks is less beneficial than 1D-CNN and 2D-CNN networks for diverse reasons. First, the data required to train these networks must be much larger which conventionally such datasets are not utilizable and methods such as pre-training, which are extensively exploited in 2D networks, cannot be used here. Another reason is that with more complicated structure of networks, it becomes much tougher to fix the number of layers, and network structure. The 3D activation map generated during the convolution of a 3D CNN is essential for analyzing data where volumetric or temporal context is crucial. This ability to analyze a series of frames or images in context has led to the use of 3D CNNs as tools for action detection and evaluation of medical imaging. \cite{a58}.

\subsubsection{Deep Belief Networks (DBNs)}

\begin{figure}[t]
    \centering
    \includegraphics[width=3.5in ]{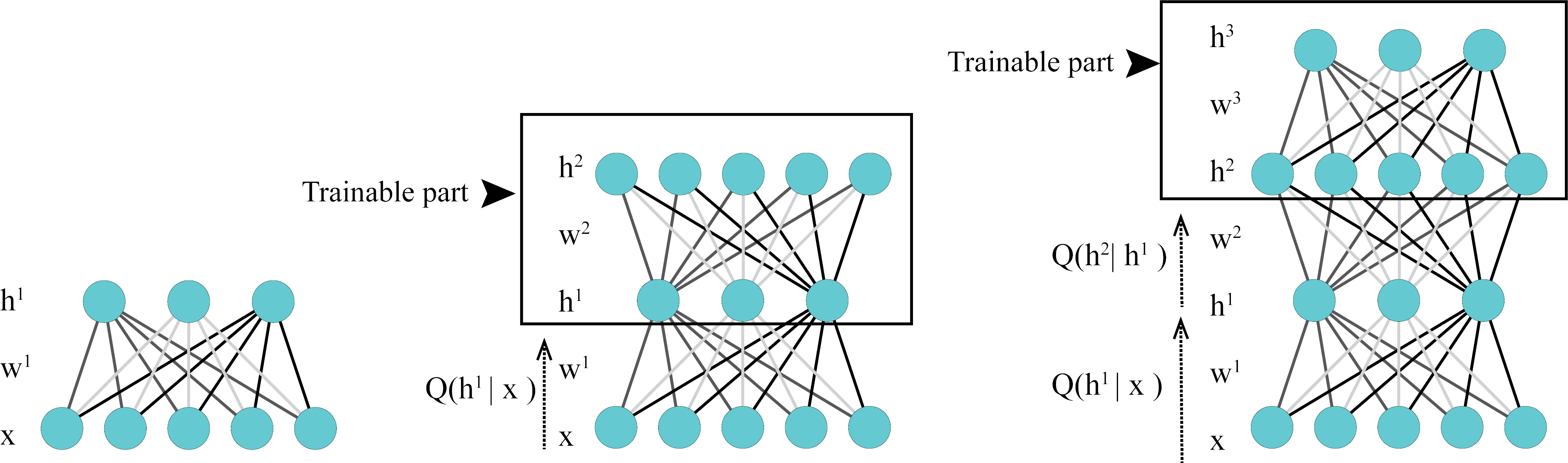}
    
    \caption{Overall block diagram of a DBN used for ASD detection.}
    \label{fig:dbs}
\end{figure}

DBNs are not  popular today as they used to be, and have been substituted by new models to perform various  applications ( e.g., autoencoders for unsupervised learning,  generative adversarial networks (GAN) for generative modes \cite{a59}, variational autoencoders (VAE) \cite{a60}). However, disregarding the restricted use of these networks in this era, their influence on the advancement of neural networks cannot be overlooked. The use of these networks in this paper is related to the feature extraction without a supervisor or pre-training of networks. These networks serve as unsupervised, consisting of several layers after the input layer, which are shown in Figure \ref{fig:dbs}. The training of these networks is done greedily and from bottom to top, in other words, each separate layer is trained and then the next layer is appended. After training, these networks are used as a feature extractor or the network weights are used as initial weights of a network for classification \cite{a21}.

\subsubsection{Autoencoders (AEs)}

\begin{figure}[t]
    \centering
    \includegraphics[width=3in ]{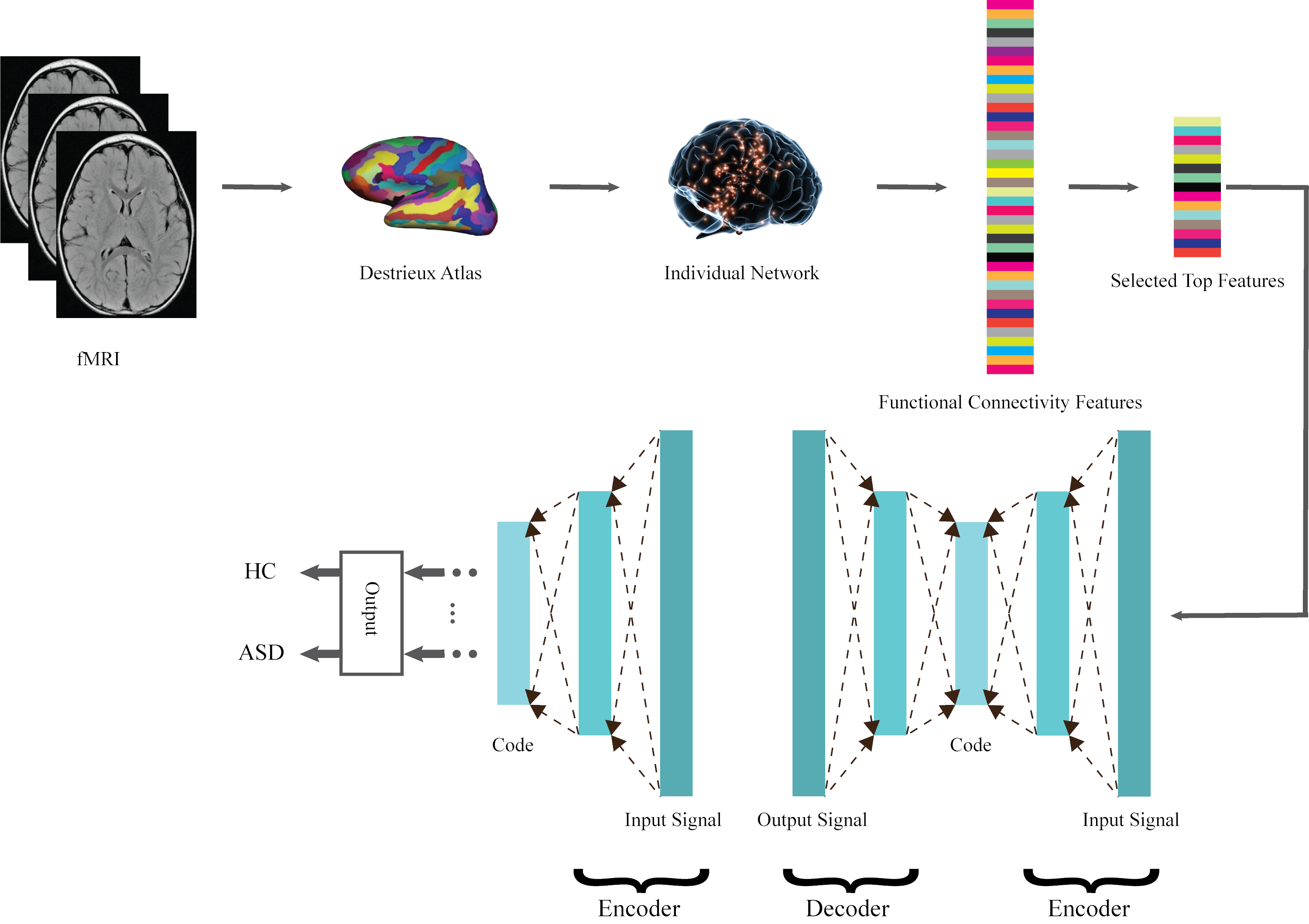}
    
    \caption{Overall block diagram of an AE used for ASD detection.}
    \label{fig:aes}
\end{figure}

Autoencoders (AEs) are more than 30 years old, and have undergone dramatic changes over the years to enhance their performance. But the overall structure of these networks has remained the same \cite{a21}.These networks consist of two parts: coder and decoder so that the first part of the input leads to coding in the latent space, and the decoder part endeavors to convert the code into preliminary data (Figure \ref{fig:aes}). Autoencoders are a special type of feedforward neural networks where the input is the same as the output. They compress the input into a lower-dimensional code and then reconstruct the output from this representation. The code is a compact “summary” or “compression” of the input, also called the latent-space representation. Various methods have been proposed to block the data memorization by the network, including sparse AE (SpAE) and denoising AE (DAE) \cite{a21}. Trained properly, the coder part of an Autoencoder can be used to extract features; creating an unsupervised feature extractor.

\subsubsection{Recurrent Neural Networks (RNNs)}
\begin{figure}[t]
    \centering
    \includegraphics[width=3in ]{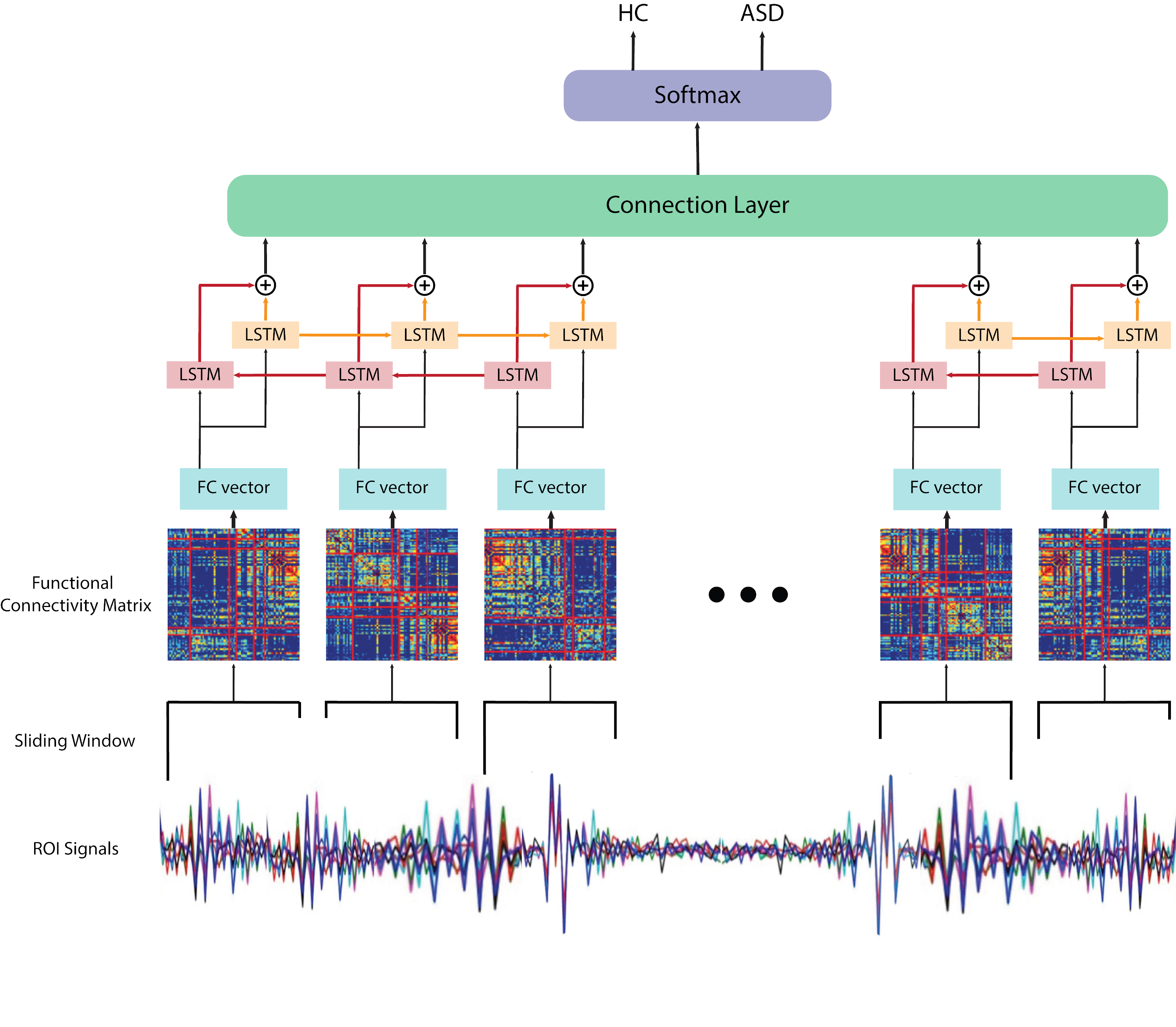}
    
    \caption{ Overall block diagram of an LSTM used for ASD detection.}
    \label{fig:rnn}
\end{figure}

In convolution networks, a kind of spatial dependencies in the data is addressed. But interdependencies between data are not confined to this model. For example in time-series, dependencies may be highly distant from each other, on the other hand, the long-term and variable length of these sequences results in that the ordinary networks do not perform well enough to process these data. To overcome these problems, RNNs can be used. LSTM structures are proposed to extract long term and short term dependencies in the data (Figure \ref{fig:rnn}). Another well-known structure called GRU is developed after LSTM, and since then, most efforts have been made to enhance these two structures and make them resistant to challenges (e.g., GRU-D \cite{a61} is used to find the lost data).

\subsubsection{CNN-RNN}
\begin{figure}[t]
    \centering
    \includegraphics[width=3.5in ]{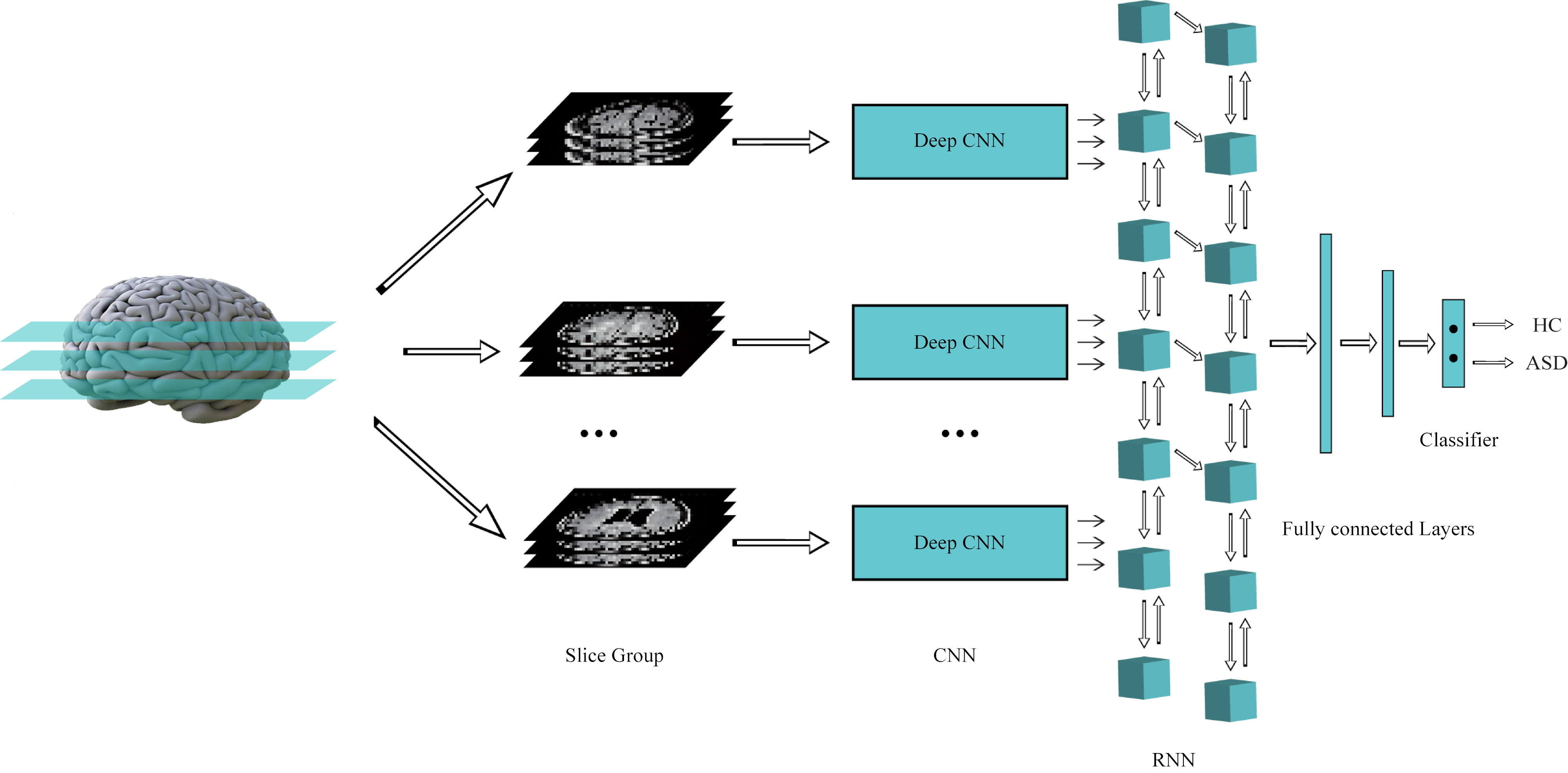}
    
    \caption{Overall block diagram of a CNN-RNN used for ASD detection.}
    \label{fig:cnnrnn}
\end{figure}

The initial idea in these networks is to utilize convolution layers to amend the performance of RNNs so that the advantages of both networks can be used; CNN-RNN, on the one hand, can find temporal dependencies with the aid of RNN, and on the other hand, it can discover spatial dependencies in data with the help of convolution layers \cite{a62}. These networks are highly beneficial for analyzing time series with more than one dimension (such as video) \cite{a63} but further to the simpler matter, these networks also yield the analysis of three-dimensional data so that instead of a more complex design of a 3D-CNN, a 2D-CNN with an RNN is occasionally used. The superiority of this model is due to the feasibility of employing pre-trained models. Figure \ref{fig:cnnrnn} demonstrates the CNN-RNN model.
\subsubsection{CNN-AE}
\begin{figure}[t]
    \centering
    \includegraphics[width=3.5in ]{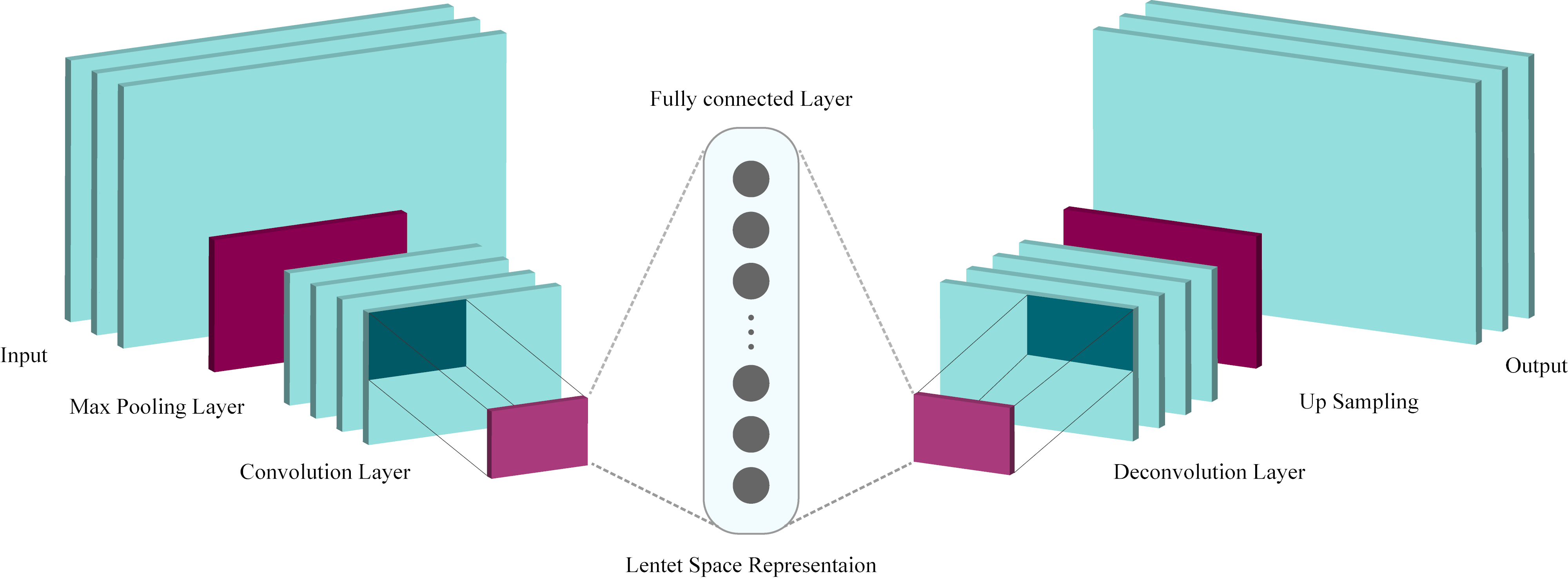}
    
    \caption{Overall block diagram of a CNN-AE used for ASD detection.}
    \label{fig:cnnae}
\end{figure}

In the construction of these networks, the principal aim and prerequisite have been to decrease the number of parameters. As shown before, changing merely the network layers to convolution markedly lessens the number of parameters; combining AE with convolution structures also makes significant contribution. This helps to exploit higher dimensional data and extracts more information from the data without changing the size of the database. Similar structures, with or without some modifications, are widely deployed for image segmentation \cite{a64}, and likewise unsupervised network can be applied for network pre-training or feature extraction. Figure \ref{fig:cnnae} depicts the CNN-AE network used for ASD detection. Tables \ref{tablelast} and \ref{table2}, provide the summary of papers published on detection and rehabilitation of ASD patients using DL, respectively.

\section{Deep Learning Techniques for ASD Rehabilitation}
Rehabilitation tools are employed in multiple fields of medicine and their main purpose is to help the patients to recover after the treatment. Various and multiple rehabilitation tools using DL algorithms have been presented. Rehabilitation tools are used to help ASD patients using mobile, computer applications, robotic devices, cloud systems, and eye tracking, which will be discussed below. Also, the summary of papers published on rehabilitation of ASD patients using DL algorithms are shown in Table \ref{table2}.
\subsection{Mobile and Software Applications }
\begin{figure}[t]
    \centering
    \includegraphics[width=3.5in ]{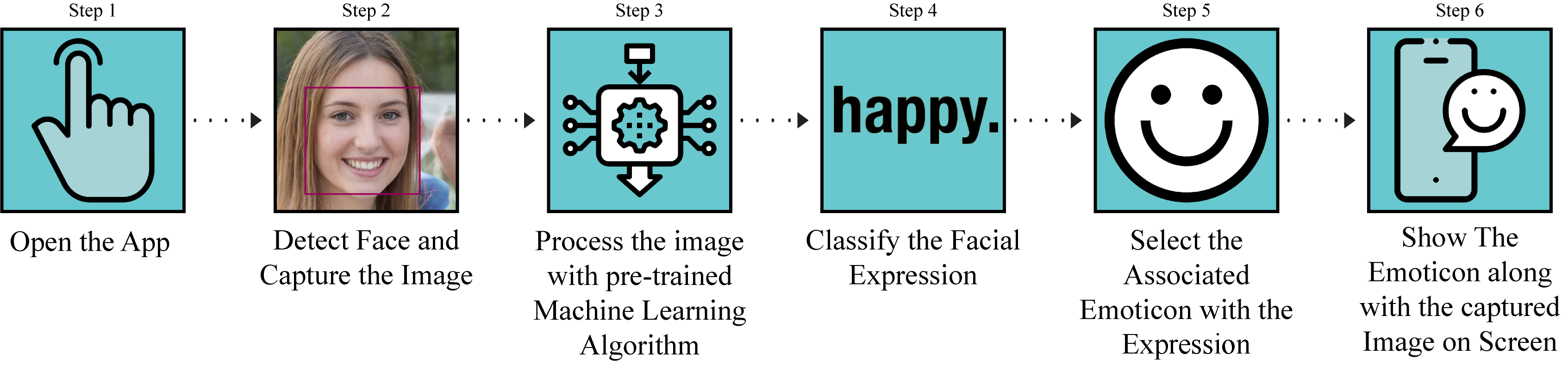}  
    
    \caption{Block diagram of ios application for ASD rehabilitation.}
    \label{fig:reiod}
\end{figure}

Facial expressions are a key mode of non-verbal communication in children with ASD and play a pivotal role in social interactions. Use of BCI systems provides insight into the user's inner-emotional state. Valles et al. \cite{a126} conducted research focused on mobile software design to provide assistance to children with ASD. They aimed to design a smart iOS app based on facial images according to Figure \ref{fig:reiod}. In this way, people's faces at different angles and brightness are first photographed, and are turned into various emoji so that the autistic child can express his/her feelings and emotions. In this group's investigation \cite{a126}, Kaggle's (The Facial Expression Recognition 2013) and KDEF (Kaggle's FER2013 and Karolinska Directed Emotional Faces) databases were used to train the VGG-16. In addition, the LEAP system was adapted to train the model at the University of Texas. The research provided the highest rate accuracy of 86.44\%. In another similar study, they achieved an accuracy of 78.32\% \cite{a124}.

\subsection{Cloud Systems}
\begin{figure}[t]
    \centering
    \includegraphics[width=2.5in ]{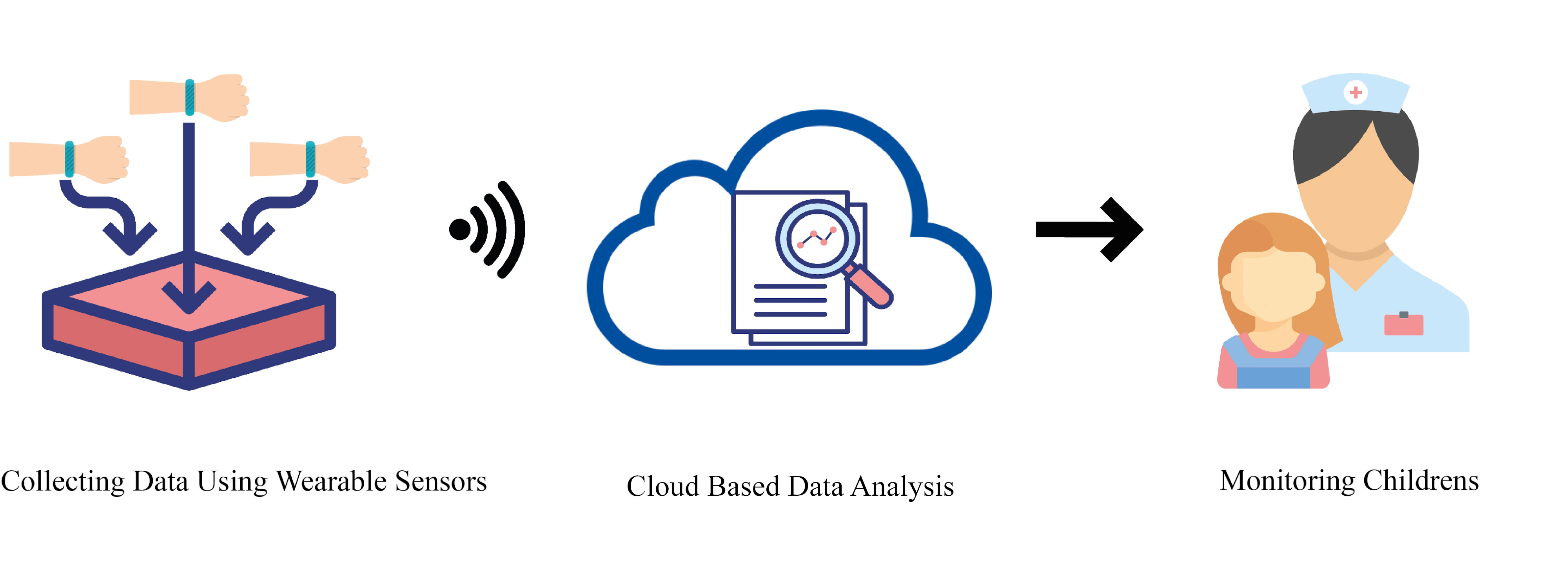}
    
    \caption{Cloud system design for ASD rehabilitation. }
    \label{fig:cl}
\end{figure}

Mohammadian et al. \cite{a143} proposed a new application of DL to facilitate automatic stereotypical motor movement (SMM) identification by applying multi-axis inertial measurement units (IMUs). They applied CNN to transform multi-sensor time series into feature space. An LSTM network was then combined with CNN to obtain the temporal patterns for SMM identification. Finally, they employed the classiﬁer selection voting approach to combine an ensemble of the best base learners. After various experiments, the superiority of their proposed procedure over other base methods was proven.  Figure \ref{fig:cl} shows the real-time SMM detection system. First, IMUs, which are wearable sensors, are used for data collection; the data can then be analyzed locally or remotely (using Wi-Fi to transfer data to tablets, cell phones, medical center servers, etc.) to identify SMMs. If abnormal movements are detected, an alarm will be sent to a therapist or parents.

\subsection{Eye Tracking}
Wu et al. \cite{a136} proposed a model of DL saliency prediction for autistic children. They used DCN in their proposed paradigm, with a SM saliency map output. The fixation density map (FDM) was then processed by the single-side clipping (SSC) to optimize the proposed loss function as a true label along with the SM saliency map. Finally, they exploited an autism eye-tracking dataset to test the model. Their proposed model outperformed other base methods. Elbattah et al. \cite{a138} aimed to combine unsupervised clustring algorithms with deep learning to help ASD rehabilitation. The first step involved the visualization of the eye-tracking path, and the images captured from this step were fed to an autoencoder to learn the features. Using autoencoder features, clustering models are developed using the K-Means algorithm. Their method performed better than other state-of-art techniques.

\onecolumn
% \begin{table*}[t]
% \vspace{-\baselineskip}
\begin{landscape}
\tiny
\begin{longtable}{|c|c|c|c|c|c|c|c|c|c|c|c|c|c|c|}
\caption{Summary of articles published using DL methods for neuroimaging-based ASD detection.}    
    \label{tablelast}\\
  \hline
  \multicolumn{1}{|c|}{\multirow{3}{*}{\shortstack[1]{Work}}} & \multicolumn{1}{c|}{\multirow{3}{*}{\shortstack[1]{Datasets}}} & \multicolumn{1}{c|}{\multirow{3}{*}{\shortstack[1]{Neuroimaging\\Modalities}}} & \multicolumn{1}{c|}{\multirow{3}{*}{\shortstack[1]{Number\\of Cases}}} & \multicolumn{1}{c|}{\multirow{3}{*}{\shortstack[1]{Pipelines}}} & \multicolumn{1}{c|}{\multirow{3}{*}{\shortstack[1]{Image\\Atlas}}} & \multicolumn{1}{c|}{\multirow{3}{*}{\shortstack[1]{Preprocessing\\Toolbox}}} & \multicolumn{1}{c|}{\multirow{3}{*}{\shortstack[1]{High level\\Preprocessing}}} & \multicolumn{1}{c|}{\multirow{3}{*}{\shortstack[1]{Inputs\\DNN}}} & \multicolumn{1}{c|}{\multirow{3}{*}{\shortstack[1]{DNN\\Toolbox}}} & \multicolumn{1}{c|}{\multirow{3}{*}{\shortstack[1]{DNN}}} & \multicolumn{1}{c|}{\multirow{3}{*}{\shortstack[1]{Number of\\Layers}}} & \multicolumn{1}{c|}{\multirow{3}{*}{\shortstack[1]{Classifier}}} & \multicolumn{1}{c|}{\multirow{3}{*}{\shortstack[1]{K fold}}} & \multicolumn{1}{c|}{\multirow{3}{*}{\shortstack[1]{Performance\\Criteria (\%)}}} \\
  \multicolumn{1}{|c|}{} & \multicolumn{1}{c|}{} & \multicolumn{1}{c|}{} & \multicolumn{1}{c|}{} & \multicolumn{1}{c|}{} & \multicolumn{1}{c|}{} & \multicolumn{1}{c|}{} & \multicolumn{1}{c|}{} & \multicolumn{1}{c|}{} & \multicolumn{1}{c|}{} & \multicolumn{1}{c|}{} & \multicolumn{1}{c|}{} & \multicolumn{1}{c|}{} & \multicolumn{1}{c|}{} & \multicolumn{1}{c|}{}\\ 
  \multicolumn{1}{|c|}{} & \multicolumn{1}{c|}{} & \multicolumn{1}{c|}{} & \multicolumn{1}{c|}{} & \multicolumn{1}{c|}{} & \multicolumn{1}{c|}{} & \multicolumn{1}{c|}{} & \multicolumn{1}{c|}{} & \multicolumn{1}{c|}{} & \multicolumn{1}{c|}{} & \multicolumn{1}{c|}{} & \multicolumn{1}{c|}{} & \multicolumn{1}{c|}{} & \multicolumn{1}{c|}{} & \multicolumn{1}{c|}{}\\
\hline
 \end{longtable}

\end{landscape}
\twocolumn

\onecolumn
% \begin{table*}[t]
% \vspace{-\baselineskip}
\begin{landscape}
\tiny
\begin{longtable}{|c|c|c|c|c|c|c|c|c|c|c|c|}
\caption{Summary of papers published on rehabilitation of ASD patients using DL algorithms.}    
    \label{table2}\\
  \hline
  \multicolumn{1}{|c|}{\multirow{3}{*}{\shortstack[1]{Work}}} & \multicolumn{1}{c|}{\multirow{3}{*}{\shortstack[1]{Datasets}}} & \multicolumn{1}{c|}{\multirow{3}{*}{\shortstack[1]{Type of\\Applications}}} & \multicolumn{1}{c|}{\multirow{3}{*}{\shortstack[1]{Number\\of Cases}}}  & \multicolumn{1}{c|}{\multirow{3}{*}{\shortstack[1]{Preprocessing}}}  & \multicolumn{1}{c|}{\multirow{3}{*}{\shortstack[1]{Inputs\\DNN}}} & \multicolumn{1}{c|}{\multirow{3}{*}{\shortstack[1]{DNN\\Toolbox}}} & \multicolumn{1}{c|}{\multirow{3}{*}{\shortstack[1]{DNNs}}} & \multicolumn{1}{c|}{\multirow{3}{*}{\shortstack[1]{Number of\\Layers}}} & \multicolumn{1}{c|}{\multirow{3}{*}{\shortstack[1]{Classifier}}} & \multicolumn{1}{c|}{\multirow{3}{*}{\shortstack[1]{K fold}}} & \multicolumn{1}{c|}{\multirow{3}{*}{\shortstack[1]{Performance\\Criteria (\%)}}} \\
  \multicolumn{1}{|c|}{} & \multicolumn{1}{c|}{} & \multicolumn{1}{c|}{} & \multicolumn{1}{c|}{} & \multicolumn{1}{c|}{} & \multicolumn{1}{c|}{} & \multicolumn{1}{c|}{} & \multicolumn{1}{c|}{} & \multicolumn{1}{c|}{} & \multicolumn{1}{c|}{} & \multicolumn{1}{c|}{} & \multicolumn{1}{c|}{}\\ 
  \multicolumn{1}{|c|}{} & \multicolumn{1}{c|}{} & \multicolumn{1}{c|}{} & \multicolumn{1}{c|}{} & \multicolumn{1}{c|}{} & \multicolumn{1}{c|}{} & \multicolumn{1}{c|}{} & \multicolumn{1}{c|}{} & \multicolumn{1}{c|}{} & \multicolumn{1}{c|}{} & \multicolumn{1}{c|}{} & \multicolumn{1}{c|}{}\\
\hline
 \end{longtable}

\end{landscape}
\twocolumn

\section{Discussion}
   
\begin{figure}[t]
    \centering
    \includegraphics[width=2.5in ]{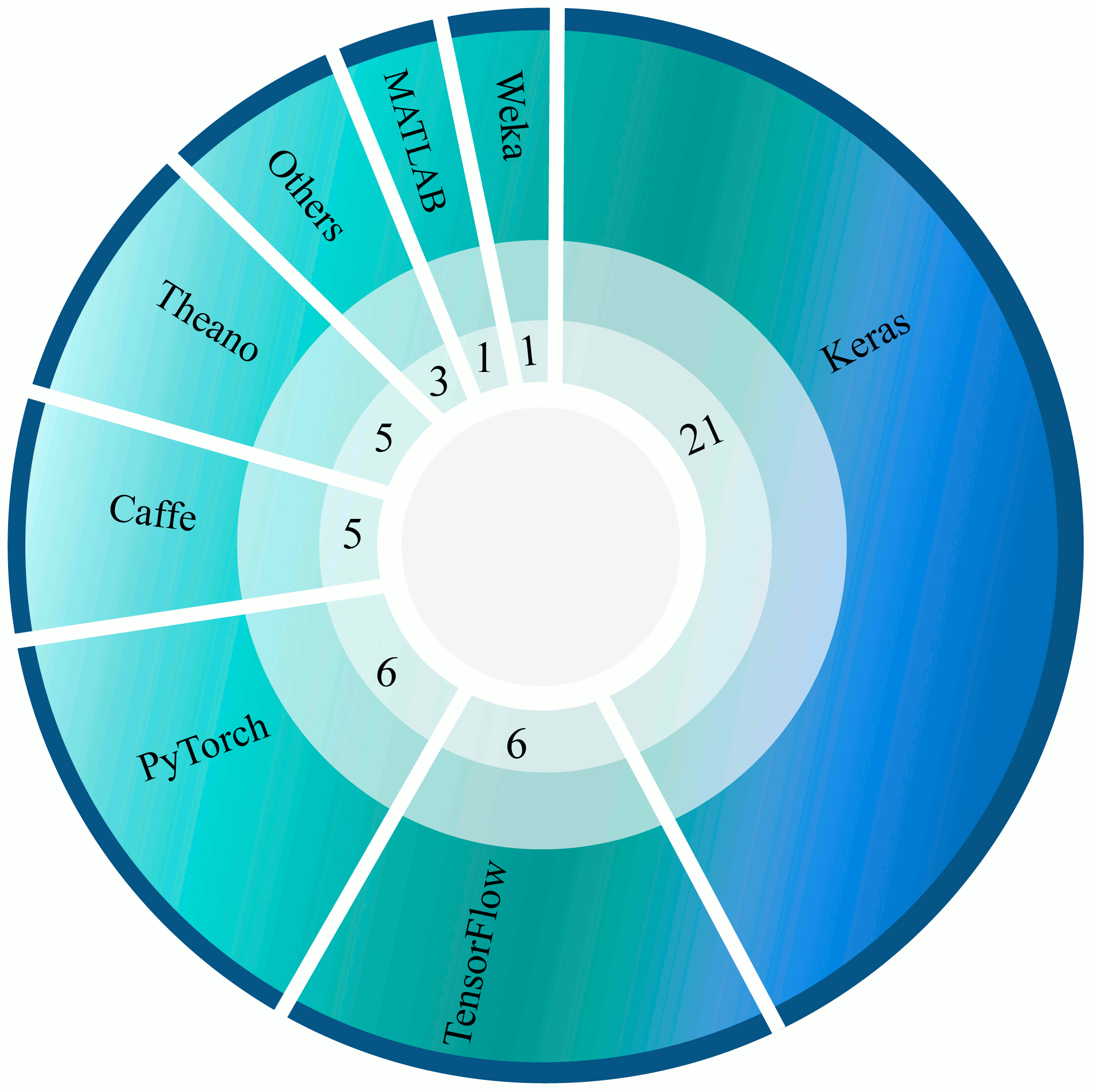}
    
    \caption{Number of DL tools used for the diagnosis and rehabilitation of ASD patients in reviewed papers.}
    \label{fig:three}
\end{figure}

In this study, we performed a comprehensive overview of the investigations conducted in the scope of ASD diagnostic CAD systems as well as DL-based rehabilitation tools for ASD patients. In the field of ASD diagnosis, numerous papers have been published using functional and structural neuroimaging data as well as rehabilitation tools, as summarized in Table \ref{tablelaster} in the appendix. A variety of DL toolboxes have been proposed for implementing deep networks. In Tables \ref{tablelast} and \ref{table2} the types of DL toolboxes utilized for each study are depicted, and the total number of their usage is demonstrated in Figure \ref{fig:three}. The Keras toolbox is used in the majority of the studies due to its simplicity. Keras offers a consistent high-level application programming interface (APIs) to build the models more straightforward, and by using powerful backends such as TensorFlow, its performance is sound. Additionally, due to all pre-trained models and available codes on platforms such as GitHub, Keras is quite popular among researchers.

\begin{figure}[t]
    \centering
    \includegraphics[width=2.5in ]{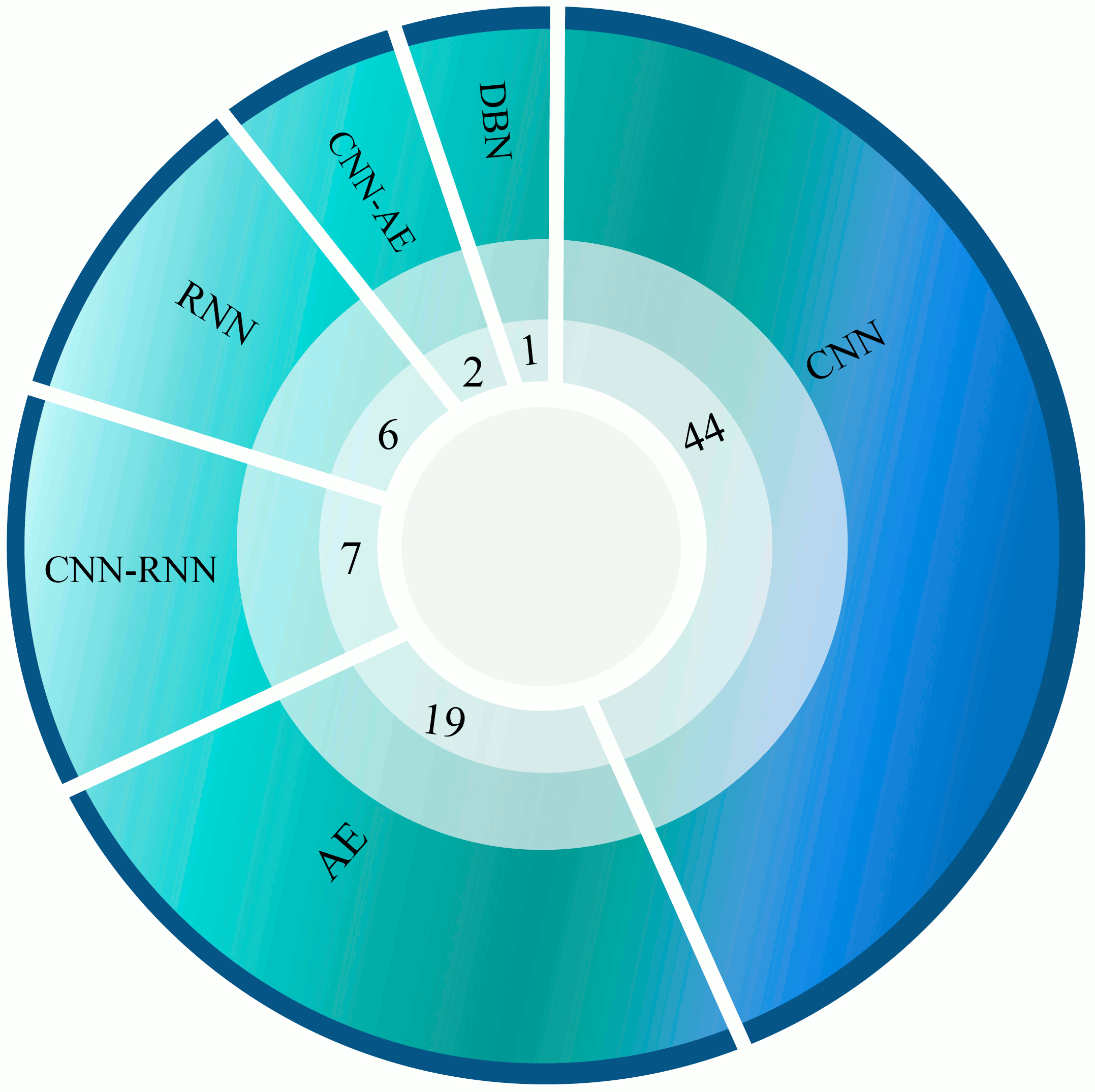}
    
    \caption{Number of of DL networks used for ASD detection and rehabilitation in the reviewed works.}
    \label{fig:four}
\end{figure}

Number of DL networks used for the ASD detection in the reviewed works is shown in Figure \ref{fig:four}.  Among the various DL architectures, CNN is found to be the most popular one as it has achieved more promising results compared to other deep methodologies.  The autoencoder, as well as RNN, have yielded favorable results. It can be noted that in recent years, the number of DL-based papers has increased exponentially due to their sound performance and also the availability of vast and thorough datasets.

\begin{figure}[t]
    \centering
    \includegraphics[width=2.5in ]{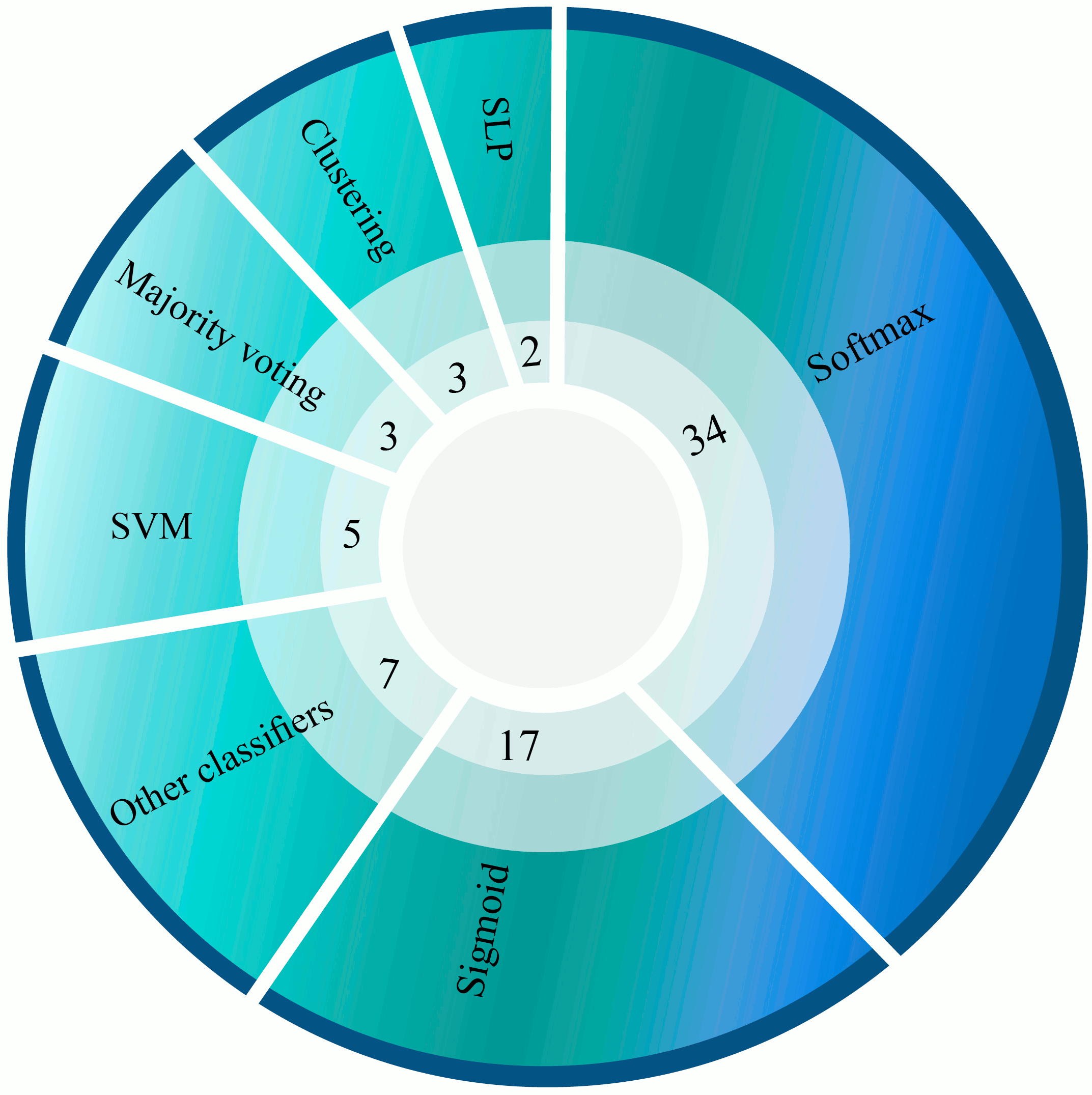}
    
    \caption{Number of various classification algorithms used for the detection of ASD and rehabilitation in DL.}
    \label{fig:five}
\end{figure}

The number of various classification algorithms used in DL networks are shown in Figure \ref{fig:five}.  One of the best and most widely used is the Softmax algorithm (Tables \ref{tablelast} and \ref{table2}). It is most popular since it is differentiable in the entire domain and computationally less expensive.

\section{Challenges}
Some of the most substantial challenges in ASD diagnosis scope using DL-based techniques are addressed in this section, which comprise database and algorithmic problems. There are only two-class brain structural and functional datasets (ASD and healthy) available in the public domain. Hence, researchers are not able to broaden their investigation to all sub-types of ASD. Two of the cheapest and most pragmatic functional neuro-screening modalities for diagnosis of ASD are EEG, and fNIRS. But unfortunately the deficiency of freely available datasets has resulted in little research in this area. Another obstacle is that multi-modality databases such as EEG-fMRI are not available to researchers to evaluate the effectiveness of incorporating information of different imaging modalities to detect ASD. Although fMRI and sMRI data are ubiquitous in the ABIDE dataset, the results of merging these structural and functional data for ASD diagnosis using DL have not yet been investigated. Another problem grappling the researchers is designing the DL-based rehabilitation systems with hardware resources. Nowadays, assistive tools such as Google Colab are available to researchers to improve the processing power; however, the problems still prevail when implementing these systems in real-world scenarios.

\section{Conclusion and Future Works}
ASD is typically characterized by social disorders, communication deficits, and stereotypical behaviors. Numerous computer-aided diagnosis systems and rehabilitation tools have been developed to assist patients with autism disorders. In this survey, research on ASD diagnosis applying DL and functional and structural neuroimaging data were first assessed. The researchers have taken advantage of deep CNNs, RNNs, AEs, and CNN-RNN networks to improve the performance of their systems. Boosting the accuracy of the system, the capability of generalizing and adapting to differing data and real-world challenges, as well as reducing the hardware power requirements to the extent that the final system can be utilized by all are the principal challenges of these systems. To enhance the accuracy and performance of CADS for ASD detection in the future, deep reinforcement networks (RL) or GANs can be exploited. Scarcity of data is always an aparamount problem in the medical field that can be resolved relatively with the help of these deep GANs. Also, as another direction for future works, handcrafted features can be extracted from data and fed to DL networks in addition to raw data; this can help increasing performance by adding the potential of traditional methods to DL-based models.

Many researchers have proposed various DL-based rehabilitation tools to aid the ASD patients.   Designing a reliable, accurate, and wearable low power consumption DL algorithm based device is the future tool for ASD patients. An achievable rehabilitation tool is to wear smart glasses to help the children with ASD. These glasses with the built-in cameras will acquire the images from the different directions of environment. Then the DL algorithm processes these images and produces meaningful images for the ASD children to better communicate with their surroundings.

% \section{}

\appendices
\section{Statistical Metrics}
This section demonstrates the equations for the calculation of each evaluation metric. In these equations, True positive (TP) is the correct classification of the positive class, True negative (TN) is the correct classification of the negative class, False positive (FP) is the incorrect prediction of the positives, False negative (FN) is the incorrect prediction of the negatives. 
\begin{equation}
    Accuracy(Acc) = \frac{TP + TN}{TP + TN + FP + FN}
\end{equation}
\begin{equation}
    Specificity(Spec) = \frac{TN}{TN + FP}
\end{equation}
\begin{equation}
    Sensitivity(Sen) = \frac{TP}{TP + FN}
\end{equation}
\begin{equation}
    Precision (Prec) = \frac{TP}{TP + FP}
\end{equation}
\begin{equation}    
    F1-Score = 2*\frac{Prec * Sens}{Prec + Sens}
\end{equation}

\textsc{Receiver Operating Characteristic Curve (ROC-Curve)}

The receiver operating characteristic curve (ROC-curve) depicts the performance of the proposed model at all classification thresholds. It is the graph of true positive rate vs. false positive rate (TPR vs. FPR). Equations for calculation of TPR and FPR are presented below.
\begin{equation}
    TPR = \frac{TP}{TP + FN}
\end{equation}
\begin{equation}
    FPR = \frac{FP}{FP+TN}
\end{equation}

\textsc{Area under the ROC Curve (AUC)}

AUC presents the area under the ROC-curve from (0, 0) to (1, 1). It provides the aggregate measure of all possible classification thresholds. AUC has a range from 0 to 1. A 100\% wrong classification will have AUC value of 0.0, while a 100\% correct classified version will have the AUC value of 1.0. It has two folded advantages. One is that it is scale-invariant, which implies how well the model is predicted rather than checking the absolute values. The second advantage is that it is classification threshold-invariant as it will verify the performance of the model irrespective of the threshold being selected.

\section{}
Table \ref{tablelaster} shows details of Deep Nets in all the papers reviewed in this study.

\section*{Acknowledgment}
MB is supported by a NHMRC Senior Principal Research Fellowship (1059660 and 1156072).
MB has received Grant/Research Support from the NIH, Cooperative Research Centre, Simons Autism Foundation, Cancer Council of Victoria, Stanley Medical Research Foundation, Medical Benefits Fund, National Health and Medical Research Council, Medical Research Futures Fund, Beyond Blue, Rotary Health, A2 milk company, Meat and Livestock Board, Woolworths, Avant and the Harry Windsor Foundation, has been a speaker for Astra Zeneca, Lundbeck, Merck, Pfizer, and served as a consultant to Allergan, Astra Zeneca, Bioadvantex, Bionomics, Collaborative Medicinal Development, Lundbeck Merck, Pfizer  and Servier - all unrelated to this work.


% \ifCLASSOPTIONcaptionsoff
%   \newpage
% \fi

\onecolumn
% \begin{table*}[t]
% \vspace{-\baselineskip}
\tiny
\begin{longtable}{|c|c|c|c|c|c|c|}
\caption{Details of Deep Nets. For ASD diagnosis and Rehabilitation.}    
    \label{tablelaster}\\
  \hline
Work	&Network	&Details for Deep Networks 	&Dropout	&Classifier	&Optimizer	&Loss Function\\
\hline
\multicolumn{1}{|c|}{\multirow{2}{*}{\shortstack[1]{\cite{a65}}}} & \multicolumn{1}{c|}{\multirow{2}{*}{\shortstack[1]{2CC3D}}} & \multicolumn{1}{c|}{\multirow{2}{*}{\shortstack[1]{CNN Layers (6) + Pooling Layers (4) + FC Layers (2)}}} & \multicolumn{1}{c|}{\multirow{1}{*}{\shortstack[1]{2 (rate=0.5)
}}} & \multicolumn{1}{c|}{\multirow{2}{*}{\shortstack[1]{Sigmoid}}} & \multicolumn{1}{c|}{\multirow{2}{*}{\shortstack[1]{NA}}} & \multicolumn{1}{c|}{\multirow{2}{*}{\shortstack[1]{BCE}}}\\
\cline{4-4}
\multicolumn{1}{|c|}{} & \multicolumn{1}{c|}{} & \multicolumn{1}{c|}{} & \multicolumn{1}{c|}{2 (rate=0.65)} & \multicolumn{1}{c|}{} & \multicolumn{1}{c|}{} & \multicolumn{1}{c|}{}\\
\hline
\cite{a66}&	2CC3D&	CNN Layers (6) + Pooling Layers (4) + FC Layers (3)&	NA&	Sigmoid&	NA&	NA\\
\hline
\cite{a67}	&3D-CNN	&CNN Layers (2) + LReLU Actication + Pooling Layers (1) + FC Layers (1)&	3 (rate=NA)	&Softmax	&SGD	&MNLL\\
\hline
\cite{a68}&	LSTM&	LSTM Layers (1) + Pooling Layers (1) + FC Layers (3)&	1 (rate=0.5)&	Sigmoid&	Adadelta&	MSE\\
\hline
\multicolumn{1}{|c|}{\multirow{2}{*}{\shortstack[1]{\cite{a69}}}} & \multicolumn{1}{c|}{\multirow{2}{*}{\shortstack[1]{CNN}}} & \multicolumn{1}{c|}{\multirow{2}{*}{\shortstack[1]{CNN Layers (2) + ReLU Activation + BN Layers (4) + FC Layers (3)}}} & \multicolumn{1}{c|}{\multirow{1}{*}{\shortstack[1]{2 (rate=0.3)}}} & \multicolumn{1}{c|}{\multirow{2}{*}{\shortstack[1]{Softmax}}} & \multicolumn{1}{c|}{\multirow{2}{*}{\shortstack[1]{Adam}}} & \multicolumn{1}{c|}{\multirow{2}{*}{\shortstack[1]{NA}}}\\
\hline

 \end{longtable}

\twocolumn

\bibliographystyle{IEEEtran}
\bibliography{main}

\end{document}